\documentclass{article}

\usepackage{arxiv}

\usepackage[utf8]{inputenc} 
\usepackage[T1]{fontenc}    
\usepackage{hyperref}       
\usepackage{url}            
\usepackage{booktabs}       
\usepackage{amsfonts}       
\usepackage{nicefrac}       
\usepackage{microtype}      
\usepackage{lipsum}
\usepackage{multicol}
\usepackage{siunitx}
\usepackage{graphicx}
\usepackage{float}
\usepackage{cite}
\usepackage{amssymb}
\usepackage{caption}
\usepackage{amsmath}

\title{Extraction of Discrete Spectra Modes from Video Data Using a Deep Convolutional Koopman Network}

\author{
  Scott B.~Leask\thanks{Corresponding author (sleask@uci.edu)} \hspace*{0.5in} Vincent G. McDonell\vspace*{0.1in}\\
  Department of Mechanical and Aerospace Engineering \\
  University of California, Irvine \\
  CA, 92697, USA \\
}

\begin{document}

\maketitle
\begin{multicols}{2}
\begin{abstract}
Recent deep learning extensions in Koopman theory have enabled compact, interpretable representations of nonlinear dynamical systems which are amenable to linear analysis. Deep Koopman networks attempt to learn the Koopman eigenfunctions which capture the coordinate transformation to globally linearize system dynamics. These eigenfunctions can be linked to underlying system modes which govern the dynamical behavior of the system. While many related techniques have demonstrated their efficacy on canonical systems and their associated state variables, in this work the system dynamics are observed optically (i.e. in video format). We demonstrate the ability of a deep convolutional Koopman network (CKN) in automatically identifying independent modes for dynamical systems with discrete spectra. Practically, this affords flexibility in system data collection as the data are easily obtainable observable variables. The learned models are able to successfully and robustly identify the underlying modes governing the system, even with a redundantly large embedding space. Modal disaggregation is encouraged using a simple masking procedure. All of the systems analyzed in this work use an identical network architecture.
\end{abstract}


\section{Introduction}

Nonlinear dynamical systems and chaotic dynamics pervade every field of study. For many practical systems, collected data are high-dimensional and a complex mix of coupled processes which exacerbate nonlinearity. Despite their prevalence, only linear systems (oftentimes with an additional constant coefficient constraint \cite{wiggins2003introduction}) can be solved analytically. This is clearly restrictive and has been met with growing efforts to model and understand systems from a data-driven perspective.

\begin{figure}[H]
  \includegraphics[width=\linewidth]{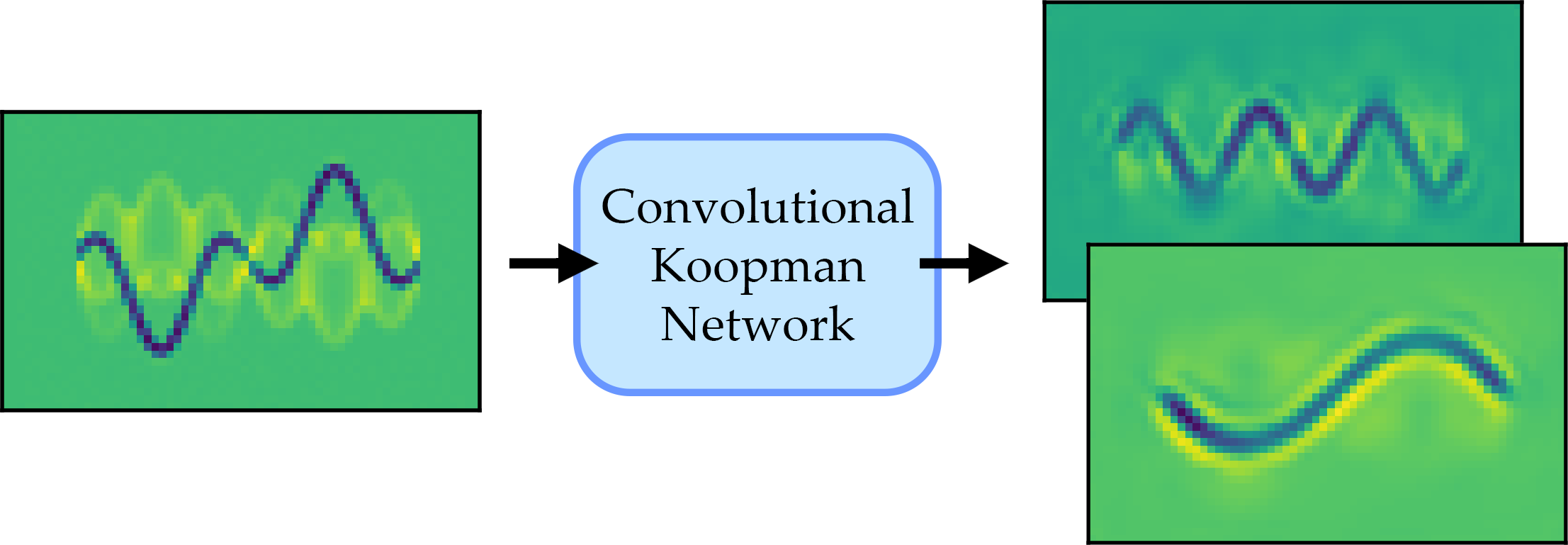}
  \vspace*{-1mm}
  \caption{Convolutional Koopman networks can perform Fourier analysis of an observed system in video format.}
  \label{fig:ckn_example}
\end{figure}

Koopman analysis has gained increasing attention recently (e.g. \cite{mezic2005spectral, williams2015data, korda2018convergence, schmid2010dynamic, tu2014dynamic}), owing to its ability to allow for linear analysis for nonlinear systems. Koopman analysis seeks to identify the Koopman operator \cite{koopman1931hamiltonian}; a possibly infinite dimensional linear map whose eigenfunctions describe the coordinate transformation to globally linearize the system dynamics. With access to the Koopman operator, nonlinear prediction, control, and system understanding can be improved.

Dynamic mode decomposition (DMD) is a widely used approach to approximate the Koopman operator which assumes the system evolves linearly \cite{schmid2010dynamic, tu2014dynamic, kutz2016dynamic}. Despite this assumption, DMD has been applied extensively to numerous problems in an attempt to extract physical understanding of a system \cite{rowley2009spectral, schmid2011applications, hua2017high, mann2016dynamic, hua2016dynamic, leask2019preliminary, proctor2015discovering, mohan2015model}. As the systems which are analyzed are typically nonlinear, DMD is inherently limited, where concerns of its application to video data have been shown in \cite{leask2019physical, leask2019use}. While Koopman analysis has the potential to improve nonlinear prediction and control, it is vital for models, and extracted modes, to be interpretable.

Significant algorithmic progress has been made in Koopman analysis, with many variants being successfully implemented \cite{towne2018spectral, chen2012variants, noack2016recursive, kutz2016multiresolution, jovanovic2014sparsity, williams2015data, li2017extended}. Recently, Koopman theory-inspired deep learning networks have demonstrated the greatest model flexibility with impressive low-dimensional representations of the Koopman operator \cite{li2017extended, yeung2019learning, lusch2018deep}. A general deep Koopman framework was presented in \cite{lusch2018deep}, which details an end-to-end trainable network capable of compactly representing dynamical systems with either discrete or continuous spectra. In this work, we naturally adapt this framework for spatio-temporal data, specifically video data, using a deep convolutional Koopman network (CKN) (Figure~\ref{fig:ckn_example}). The CKN supports 2D and 3D convolutions with minimal alteration from the original deep Koopman framework, owing to its autoencoder design.

Focus on video data is motivated by the practicality of collecting optical-based measurements of a system, which can be done non-intrusively, capturing a large spatial extent with high spatio-temporal resolution.

\section{Convolutional Koopman Network}

A deep Koopman network attempts to learn a low-dimensional approximation of the Koopman operator. Koopman theory states that, under an appropriate coordinate transformation, a nonlinear dynamical system can be globally linearized, with its dynamics captured by a possibly infinite-dimensional linear map, $\mathcal{K}$, the Koopman operator.

Consider a discrete-time system
\begin{equation}
    \mathbf{x}_{k+1} = \mathbf{f}(\mathbf{x}_k)
\end{equation}
where $\mathbf{f}$ is a nonlinear function of the state variables, $\mathbf{x}\in\mathbb{C}^{n}$. Commonly, one has access to the state of observable variables instead, which itself is a nonlinear function of $\mathbf{x}$. The goal is to learn a transformation, $\phi$, and a linear operator, $\mathcal{K}$, such that
\begin{equation}
    \phi(\mathbf{x}_{k+1}) = \mathcal{K}\phi(\mathbf{x}_{k}).
    \label{eq:Koop_eq}
\end{equation}
Deep Koopman networks simultaneously learn $\phi$ and a low-dimensional representation of $\mathcal{K}$, here given by $\mathbf{K}(\lambda)$. The general form of $\mathbf{K}(\lambda)$ is a Jordan matrix, whose blocks represent independent system modes characterized by a pair of complex conjugate eigenvalues,~$\lambda$. We can visualize the effect of each eigenvalue pair independently, e.g. by nulling the contribution of all modes except the mode of interest.

\subsection{Architecture}

\begin{figure}[H]
  \vspace*{0.1in}
  \includegraphics[width=\linewidth]{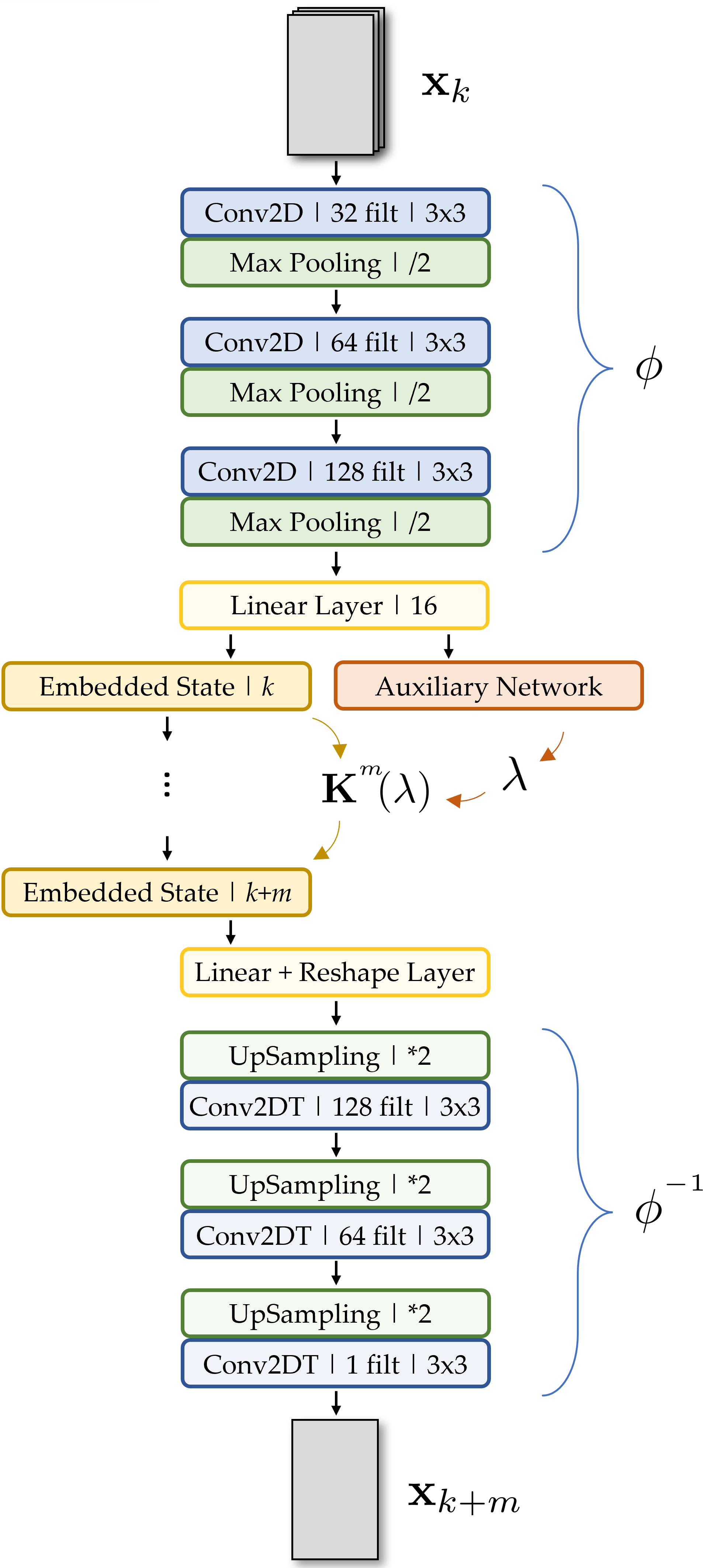}
  \vspace*{0.1in}
  \caption{The CKN architecture used in this work. $3\times 3$ filters were used in all convolutional layers, up-sampling layers were used to act as inverse to max pooling, and reshape layers were used to ensure shape compatibility between convolutional and linear layers. The embedded state is mapped forward in time by $\mathbf{K}(\lambda)$, which is parameterized by the auxiliary network.}
  \label{fig:ckn_arch}
  \vspace*{0.1in}
\end{figure}

The deep convolutional Koopman network (CKN) architecture naturally adapts the Koopman framework provided in \cite{lusch2018deep} for video data. The CKN used in this work is shown in Figure~\ref{fig:ckn_arch}. The encoder and decoder sections are composed of convolutional and pooling layers, and their transposes, respectively, which learn the coordinate transformation, $\phi$ in (\ref{eq:Koop_eq}). While it is easy to extend this to 3D convolutional layers, no benefit was found over the 2D convolutional networks. Indeed, different approaches are available in convolving together spatio-temporal information which leads to various impacts on model performance \cite{tran2015learning, tran2018closer}.

Batch normalization \cite{ioffe2015batch} and dropout \cite{srivastava2014dropout} are applied after each layer, with a dropout rate of 0.75. The last layer of the auxiliary network and the decoder and the two linear layers use linear activation functions, all other layers use the ReLU activation function \cite{nair2010rectified}. Models were trained using TensorFlow 2.0 \cite{abadi2016tensorflow}.

The auxiliary network learns the eigenvalues which parameterize an approximation of the Koopman operator, $\mathbf{K}(\lambda)$. This auxiliary network allows both discrete and continuous spectra to be identified, and was kept for this study. Generally, $\mathbf{K}(\lambda)$ is applied $m$-times before being decoded back into the original state space. Max pooling is non-invertible, so the decoder approximates the inverse of $\phi$ through up-sampling and transpose convolutions \cite{zeiler2014visualizing}.

The exact same architecture is used for each system. Three convolutional layers, each followed by a max pooling layer with pool size 2, make up the encoder, with the transpose of each layer, in reverse order, making up the decoder. 8 pairs of complex conjugate eigenvalues were used to parameterize the Koopman operator. Although only systems with discrete spectra are analyzed in this work, we keep the auxiliary network of \cite{lusch2018deep}. In practice, the model results are the same, but the training time required to converge is increased to maintain generality.

For modal extraction, all data are used when training. Other modal decomposition techniques, such as proper orthogonal decomposition or DMD, for example, use all of the collected data to identify system modes. Using the CKN architecture primarily for future state prediction, for example, warrants the use of validation and test data.

Though not strictly necessary, the mean of the data were first subtracted as this was found to improve convergence and modal extraction robustness. Mean subtraction is not advised for DMD \cite{chen2012variants}, and so the effect on Koopman networks is of future interest.

\subsection{Loss Function}

The overall loss function we use is:
\begin{equation}
\begin{split}
    \mathcal{L} = \hspace*{2mm} & \lVert \mathbf{x}_1 - \phi^{-1}(\phi(\mathbf{x}_1))\rVert \\
    & + \frac{1}{T}\sum_{m=1}^{T}\lVert \mathbf{x}_{m+1} - \phi^{-1}(\mathbf{K}^{m}\phi(\mathbf{x}_1))\rVert \\
    & + \frac{1}{T}\sum_{m=1}^{T}\lVert \phi(\mathbf{x}_{m+1}) - \mathbf{K}^{m}\phi(\mathbf{x}_1)\rVert \\
\end{split}
\end{equation}
where all norms are the mean squared error, and $T$ is the number of time steps the network predicts into the future. Perhaps surprisingly, large values of $T$ suffered more easily from local saddle points or minima, and so $T=3$ was used for all systems.

\subsection{Input Masking}

Without alteration, the CKN will learn efficient Koopman embeddings which allow for accurate nonlinear prediction. However, there is no guarantee that extracted modes are interpretable. In initial experiments, the network was able to learn embeddings which were \textit{too} compact, whereby interpretation was inhibited. To overcome this, a simple input masking scheme was implemented which robustly identifies system modes and improves interpretability.

Input images are masked with a random number of, possibly overlapping, squares of random side length. While the random values likely depends on the systems of interest, 2 or 3 squares with a side length between 10\%-30\% of the image width or height worked well for this study. The intuition behind the effectiveness of this approach is that the masking blocks the contribution of certain modes, thus allowing the network to identify independent processes.

\section{Results}

We demonstrate the capabilities of the CKN with simple, simulated systems in order to verify the efficacy of the extracted modes. These systems have discrete spectra and each is composed of multiple distinct modes, which may overlap, with known spatio-temporal coherence. The goal of the CKN is to successfully disaggregate the modes. The time-step between each successive frame is \SI{0.01}{\second}.

Since the size of the embedding space is larger than the number of true underlying modes, numerous modes are redundant. Indeed, results can be refined by tuning the embedding space for a given problem, however, this was not done to reduce manual tuning. The selection of valid modes was straightforward as redundant mode frequencies tend to group around true system modes but with indistinct structures or are aperiodic. A comparison of the frequencies of the extracted modes with the ground truth frequencies for the systems analyzed are given in Table~\ref{tab:freqs}.

We note that the network is still successful when applied to nonlinear, continuous spectra systems. Future work will focus on those systems, however, example results for the nonlinear pendulum are given in the appendix.

\begin{table*}[t]
  \centering
  \bgroup
    \def\arraystretch{1.5}
    \vspace*{5mm}
    \begin{tabular}{p{3cm} p{5cm} p{5cm}}
     \hline
     System & Ground Truth & CKN Extracted Frequencies\\
     \hline
     \hline
     Linear Waves & (\SI{2.00}{\hertz}, \SI{5.71}{\hertz}, \SI{5.00}{\hertz}) & (\SI{1.99}{\hertz}, \SI{5.71}{\hertz}, \SI{4.93}{\hertz}) \\
     Particles & (\SI{5.89}{\hertz}, \SI{7.84}{\hertz}, \SI{3.92}{\hertz}) & (\SI{5.71}{\hertz}, \SI{7.35}{\hertz}, \SI{3.91}{\hertz}) \\
     Nonlinear Wave & (\SI{1.78}{\hertz}, \SI{2.67}{\hertz}) & (\SI{1.76}{\hertz}, \SI{2.63}{\hertz}) \\
     \hline
    \end{tabular}
    \egroup
    \vspace*{3mm}
    \caption{Frequencies of extracted modes compared to the ground truth. The frequencies are tuples ordered by mode number.}
  \label{tab:freqs}
\end{table*}

\subsection{Overlapping Waves}

Generally, system processes overlap or occlude other processes and it is visually unclear what the governing processes are. The CKN is tasked with a simple wave problem: given a top-down view of numerous waves emanating from different sources, extract the independent sources' location, frequency, wavelength, and decay rate. Each pixel directly captures the amplitude dynamics at that spatial location.

\begin{figure*}
  \centering
  \includegraphics[width=0.8\textwidth]{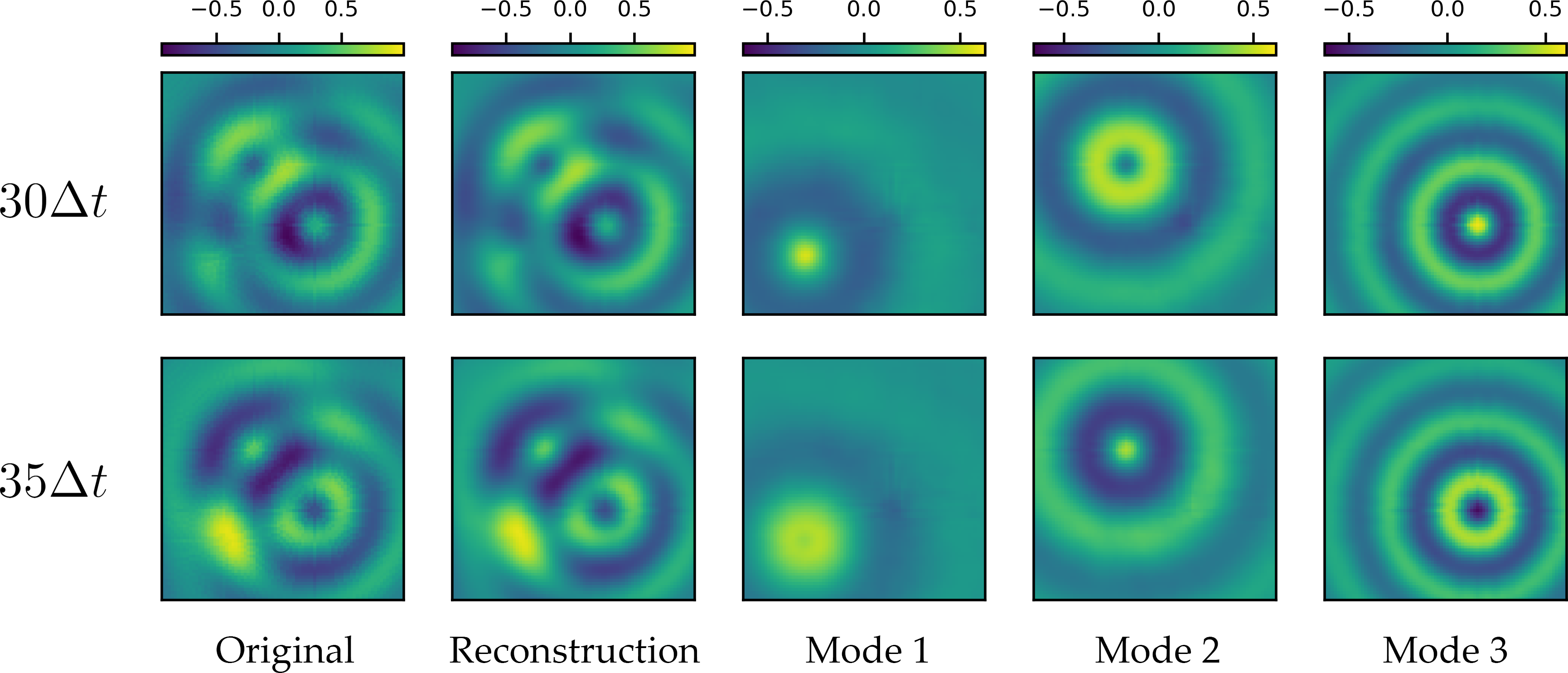}
  \vspace*{2mm}
  \caption{\textbf{\textit{Linear Waves.}} All three superimposed waveforms, with their location, frequency, wavelength, and decay rate, are successfully identified by the network.}
  \label{fig:wave_result}
\end{figure*}

The results of the modal decomposition are given in Figure~\ref{fig:wave_result}, predicting 30 and 35 time steps into the future. We choose to visualize the modes as partial state reconstructions as it intuitively demonstrates the direct contribution of each mode. The CKN is able to accurately isolate each independent source and its accompanying parameters. Even with a redundantly large embedding space, the network has no problem disaggregating the modes.

\subsection{Constant Velocity Particles}

A simple case where DMD fails to compactly capture system modes for video data is for a group of observed particles moving on linear trajectories at constant velocity \cite{leask2019use}. Three particles move at different velocities and thus have different periods as they wrap in and out of the frame (i.e. motion is top to bottom). One of the particle's frequency is a harmonic of another.

\begin{figure*}
  \centering
  \includegraphics[width=0.8\textwidth]{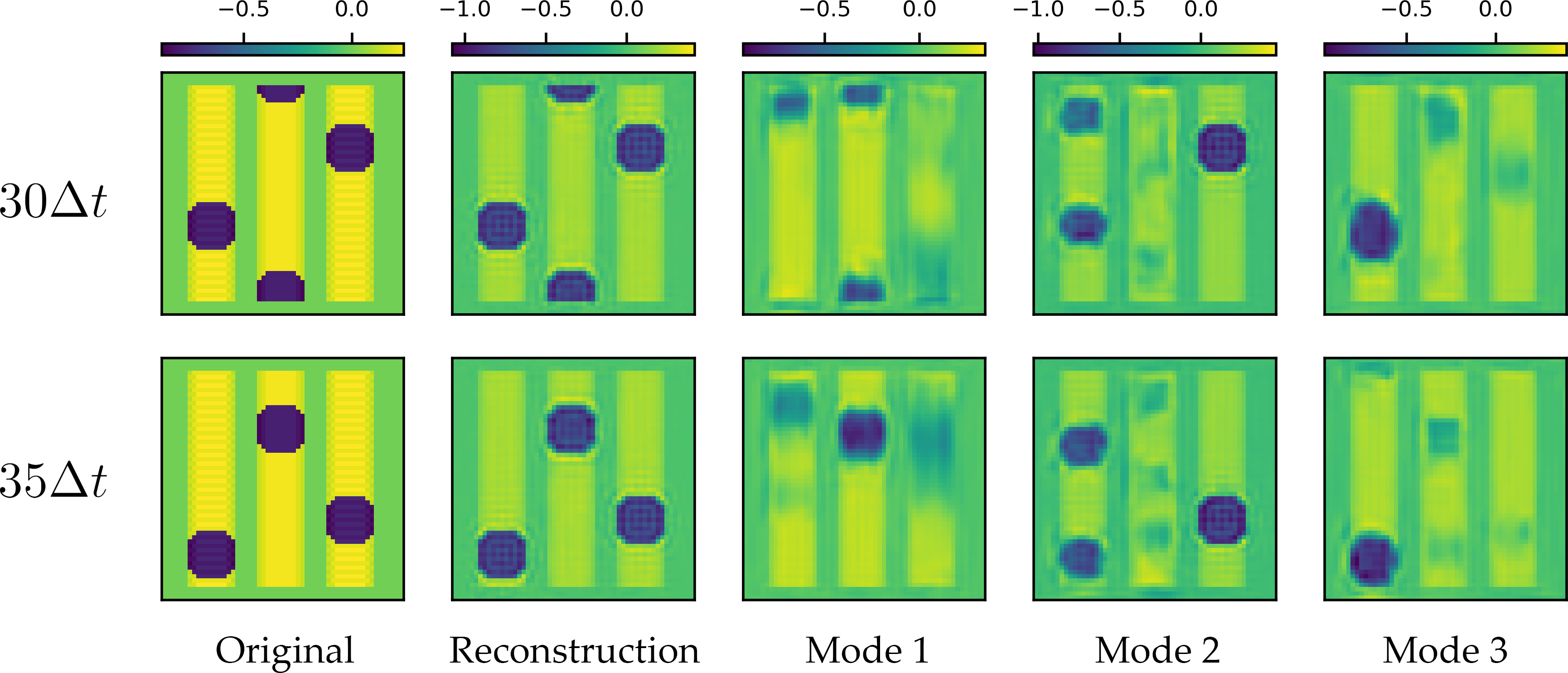}
  \vspace*{2mm}
  \caption{\textbf{\textit{Particles.}} The network is able to extract the three underlying modal frequencies (Table~\ref{tab:freqs}) and the main spatio-temporal contribution of each associated process. Mode 2 captures behaviors of its corresponding fundamental mode, mode 3.}
  \label{fig:particle_result}
\end{figure*}

The time evolution of the input data, a three-mode reconstruction, and the contributions of each mode are given in Figure~\ref{fig:particle_result}. The network is able to disaggregate the distinct particle behaviors, although some artifacts are present. Contributions of mode 3 is captured by mode 2, its second harmonic, but at a weaker intensity and with double the number of particles. In this case, it is easy to associate cross-modal coupling between mode 2 and mode 3. In general, this association will not be as evident, so it is of great interest to fully disaggregate modes which happen to be at harmonic frequencies.

Other artifacts arise due to contributions of higher-order modes (i.e. mode 4+). The frequencies of these higher-order modes group around the shown modes, but whose spatio-temporal structures are not distinct. Ideally, the network can learn to null redundant modes.

The masking procedure was found to be critical for this system. Without allowing the network to focus on specific spatio-temporal regions, the CKN would group independent processes. This was particularly the case for modes 2 and 3, where only a single mode was actually needed to capture both particles.

\begin{figure*}
  \centering
  \vspace*{5mm}
  \includegraphics[width=0.8\textwidth]{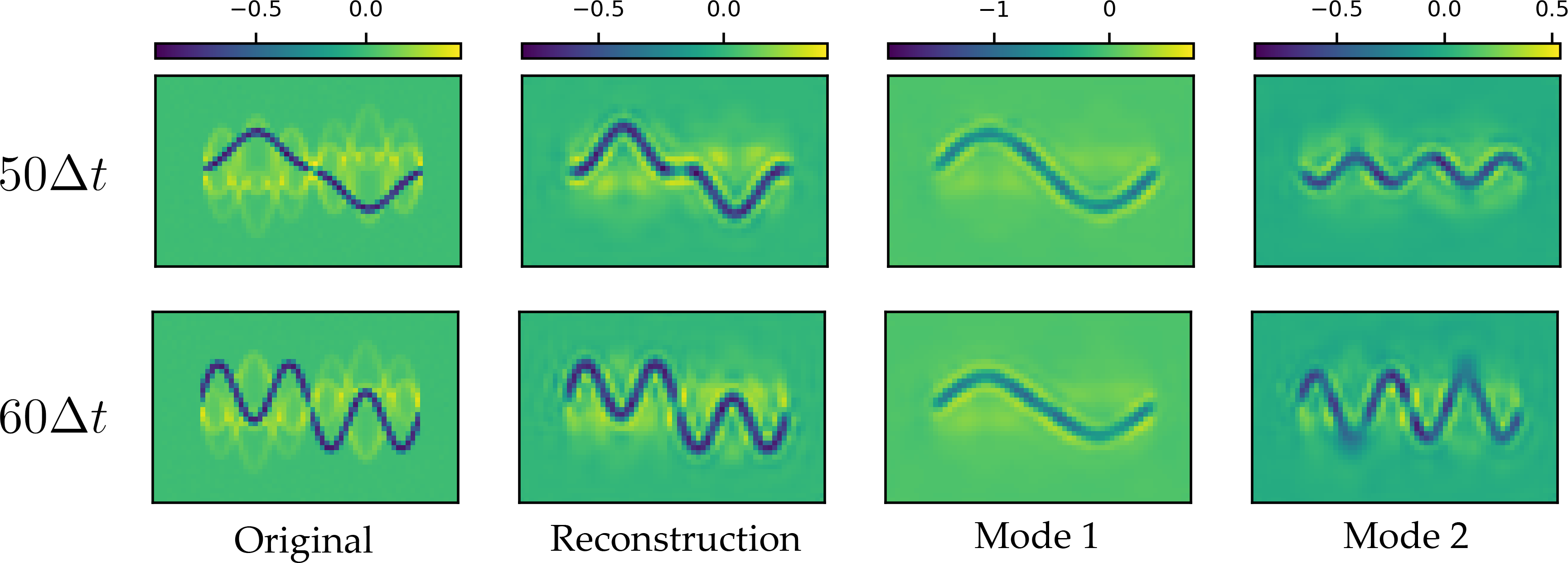}
  \vspace*{2mm}
  \caption{\textbf{\textit{Nonlinear Wave.}} Given a nonlinear representation of a superimposed waveform, the CKN learns the nonlinear transformation to linearize the dynamics. The extracted modes are the correct spatio-temporal structures of the constituent sinusoids.}
  \label{fig:string_result}
\end{figure*}

\subsection{Visual Fourier Analysis}

The previous wave system has the benefit of access to the intrinsic state variables of the waves. A more interesting problem is: can an observed periodic waveform, without direct access to amplitude measurements, be decomposed into a sum of sinusoids? In this system, a string vibrates according to the superposition of two sinusoids and the data are recorded observing the changes in amplitude on the y-axis. This change in perspective for a Fourier problem makes the problem nonlinear and DMD is unable to compactly capture the two modes.

To avoid ambiguity in the direction of velocity, three consecutive frames, concatenated together in the channel dimension, were used as inputs to the CKN.

The results of the modal decomposition are given in Figure~\ref{fig:string_result}. Without access to the intrinsic amplitude measurements, the network is still able to successfully decompose the waveform into its constituent sinusoids. The network has learned to associate pixel height with wave amplitude; the reconstructed wave amplitudes (\mbox{y-axis} displacements) are a linear combination of the extracted modal wave amplitudes.

Where DMD may fail due to the observables analyzed, the convolutional Koopman network is able to learn the suitable nonlinear transformation to linearize the dynamics.

\section{Conclusion}

In this paper, we demonstrate the capabilities of a deep convolutional Koopman network (CKN); a deep learning-extension of Koopman analysis specifically for image and video data. Due to the ubiquity and practicality of optical-based measurements, video data of simple dynamical systems have been analyzed and their underlying processes can be extracted with the CKN architecture.

Three systems with discrete spectra were analyzed with known modes to serve as ground truth data. The CKN is flexible; accurate extracted modes were obtained even with a redundantly large embedding space and parameterizing the Koopman matrix with an auxiliary network. Of course, training can be improved and the number of modal artifacts may be reduced by eliminating the contributions of extraneous modes, however, this needs to be investigated further.

A main motivation for the CKN is in overcoming limitations of dynamic mode decomposition (DMD) when applied to nonlinear observable data. The CKN can perform a visual Fourier decomposition by first learning the nonlinear transformation to linearize the dynamics which makes Fourier analysis trivial.

Although only systems with discrete spectra were analyzed here, the CKN can still be applied to continuous spectra systems (see the appendix for preliminary results). However, the discrete spectra modes highlight a number of important observations: (1) a masking procedure greatly aids in the disaggregation of modes, (2) harmonic modes have weak coupling with fundamental modes, and (3) a large embedding space offers extraction flexibility at the cost of modal artifacts.

\bibliographystyle{ieeetr}  
\bibliography{references}

\section*{Appendix}
\subsection*{Nonlinear Pendulum}

\begin{figure}[H]
  \includegraphics[width=\linewidth]{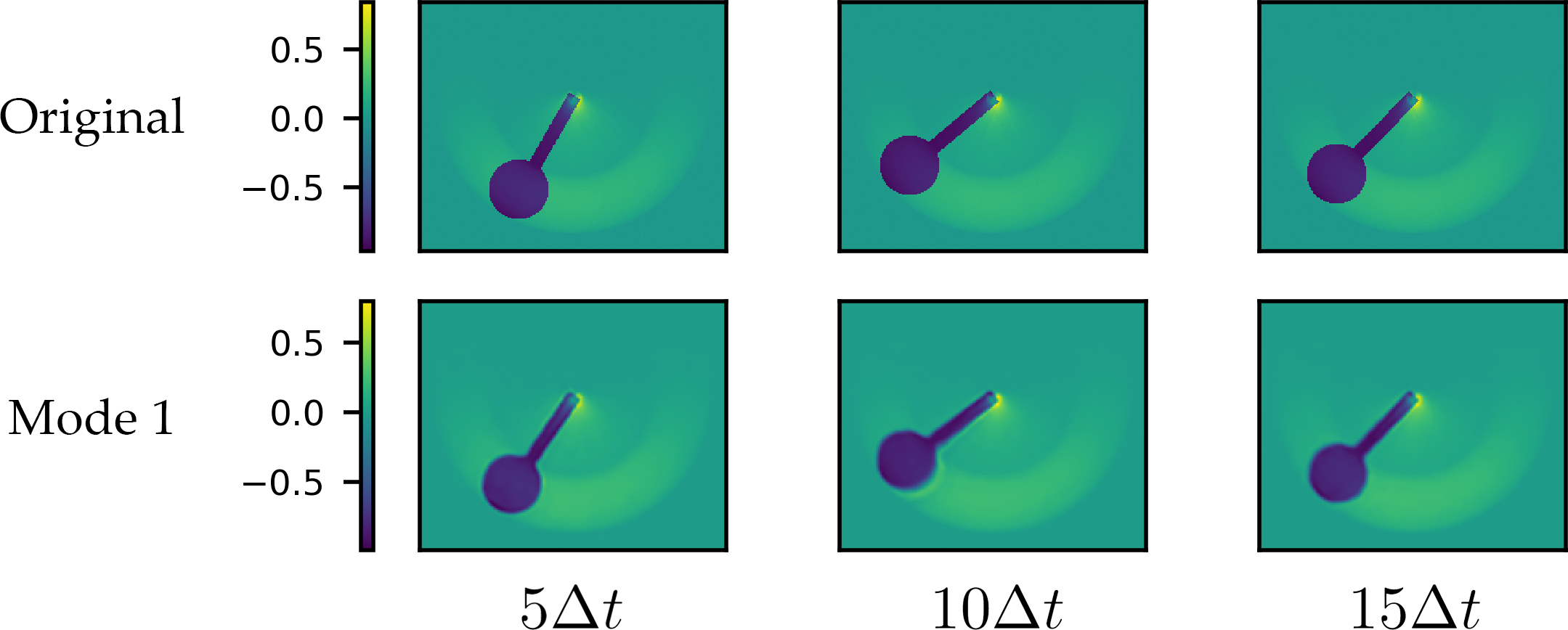}
  \vspace*{1mm}
  \caption{Comparison of original data with future prediction using only a single mode. The CKN successfully captures the correct period and change in pendulum velocity.}
  \label{fig:pend_result}
\end{figure}

The convolutional Koopman architecture performs well for nonlinear systems with continuous spectra, such as the nonlinear pendulum example, with results shown in Figure~\ref{fig:pend_result}. Using just a single mode, parameterized by the auxiliary network, the CKN correctly captures the continuous spectra, i.e. different periods.

To avoid ambiguity in the direction of velocity, three consecutive frames, concatenated together in the channel dimension, were used as inputs to the CKN. The network was trained 5 time steps into the future from the last frame used in the input. To successfully capture the pendulum dynamics, the spatial resolution must increase to allow the network to discern changes in angle and angular velocity. Of course, this increases the burden on computational resources, so is left to future studies to investigate more completely. The data resolution for the results in Figure~\ref{fig:pend_result} is $208\times256$ pixels.

For discrete spectra modes, visualizing modes can be done by creating a modal video of the spatio-temporal behaviors captured. For continuous spectra modes, where an infinite number of varying periods can be captured, this same visualization does not hold. Identifying interpretable and useful \textit{representations} of continuous spectra modes is of great interest.

\end{multicols}
\end{document}